\documentstyle[eaclap]{article}

\author{Manny Rayner \\
SRI International \\
Suite 23, Millers Yard\\
Cambridge CB2 1RQ, United Kingdom \\
{\tt manny@cam.sri.com}
\And
Pierrette Bouillon \\
ISSCO, University of Geneva\\
54, route des Acacias\\
1227 Geneva, Switzerland\\
{\tt pb@divsun.unige.ch}}
\title{Hybrid Transfer in an\\
       English-French Spoken Language Translator}

\begin{document}

\maketitle
\bibliographystyle{acl}

\begin{abstract}

The paper argues the importance of high-quality translation for
spoken language translation systems. It describes an
architecture suitable for rapid development of high-quality
limited-domain translation systems, which has been implemented within
an advanced prototype English to French spoken language translator.
The focus of the paper is the hybrid transfer model which combines
unification-based rules and a set of trainable statistical
preferences; roughly, rules encode domain-independent grammatical
information and preferences encode domain-dependent distributional
information.  The preferences are trained from sets of examples
produced by the system, which have been annotated by human judges as
correct or incorrect.  An experiment is described in which the model
was tested on a 2000 utterance sample of previously unseen data.

\end{abstract}

\section{Introduction}\label{Section:Introduction}

During the last five years, people have started to believe there is a serious
possibility of building practically useful spoken language translators for
limited domains.  There are now a number of high-profile projects with large
budgets, the most well-known being the German Verbmobil effort.  At the moment,
the best systems are at the level of advanced prototypes; making projections
from current performance, it seems reasonable to hope that these could be
developed into commercially interesting systems within a time-scale of five to
ten more years.

This paper will describe work carried out on one such advanced
prototype, the Spoken Language Translator (SLT) system
\cite{SLT-HLT,SLT-report}. SLT can translate spoken English
utterances from the domain of air travel planning (ATIS; \cite{ATIS})
into spoken Swedish or French, using a vocabulary of about 1200 stem
entries. The Swedish version has been operational since June 1993, and
has been publicly demonstrated on numerous occasions. The French
version became operational fairly recently; the language-processing
component was demoed for the first time at the CeBIT trade fair at
Hannover in March 1995. The Swedish and French versions have
approximately equivalent levels of performance \cite{SLT-ICSLP}. SLT
incorporates modules for speech recognition, speech synthesis and
translation. In the paper, we will focus on the last of these. All
examples given will refer to the French version.

One of the most important differences between spoken language
translation and text translation is that there are much stronger
demands on quality of output.  If output is not good enough, people
frequently have difficulty understanding what has been said. There is
no possibility of the pre- or post-editing which nearly all text
translation systems rely on.  Quite apart from the problem of generating
natural-sounding speech, it is also necessary to ensure that the
translated text sent to the speech synthesizer is itself of sufficient
quality.  A high-quality translation must fulfill several criteria: in
particular, it should preserve the meaning of the original utterance,
be grammatical, and contain correct word-choices.

The basic design philosophy of the SLT project has been to build a framework
which
is theoretically clean, on the usual grounds that this makes for a
system that is portable and easy to scale up.  We have attempted to subsume as
much of the system as possible under two standard paradigms:  {\it
unification-based language processing} and the {\it noisy-channel statistical
model}.  The unification-based part of the system encodes domain-independent
grammatical rules; for each source-language word or grammatical construction
covered by the system, it describes the possible target-language translations.
When the rules permit more than one potentially valid translation, the
statistical component is used to rank them in order of relative plausibility.
The next two paragraphs give some examples to motivate this division of
knowledge sources.

The simplest examples of transfer rules are those used to translate individual
words; here it is immediately clear that many words can be translated in
several ways, and thus that more than one rule will often apply.  For
instance, the English preposition {\it ``on''} can be translated as any of the
French prepositions {\it ``avec''} ({\it fly to Boston on Delta $\rightarrow$
aller \`{a} Boston avec Delta}); {\it ``sur''} ({\it information on ground
transportation $\rightarrow$ des renseignements sur les transports publics});
{\it ``\`{a} bord de''} ({\it a meal on that flight $\rightarrow$ un repas
\`{a} bord de ce vol}); {\it ``pour''} ({\it the aircraft which is used on
this flight $\rightarrow$ l'avion qu'on utilise pour ce vol}); or omitted and
replaced by an implicit temporal adverbial marker ({\it leave on Monday
$\rightarrow$ partir le lundi}).  In each of these cases, the correct choice
of translation is determined by the context.

To take a slightly more complex case, which involves some grammar, there are
a number of transfer rules that list possible ways of realizing the English
compound nominal construction in French.  Among these are adjective + noun
({\it economy flight} $\rightarrow$ {\it vol \'{e}conomique}); noun + PP
({\it arrival time} $\rightarrow$ {\it heure d'arriv\'{e}e}; {\it
Boston ground transportation} $\rightarrow$ {\it transports publics
\`{a} Boston}); or in special cases simply a compound noun ({\it Monday
morning} $\rightarrow$ {\it lundi matin}).  Again, the individual lexical
items and the context determine the correct rule to use.

Experience has shown that it is relatively simple to write the
context-independent rules which list sets of choices like the ones
above.  It is however much more difficult to use rules to specify the
context in which each particular choice is appropriate.  Moreover, the
correct choice is frequently domain-dependent; thus the rules will
need to be rewritten if the system is ported to a new application.
For these reasons, statistically trained machine translation
architectures have recently been receiving a great deal of attention.
Some researchers (notably those in the IBM CANDIDE project,
\cite{Brown:90}) have even gone so far as to claim that statistical
techniques are sufficient on their own.  Our view is that this is at
best unnecessary. Since many aspects of language (for instance,
agreement and question-formation in French) appear to be regular and
readily describable by rules, it seems more logical to use a mixture
of rules and statistics; it is in this sense that we have a {\it
hybrid} transfer model
(cf. \cite{Brown:92,Carbonell:92,GrishmanKosaka:92}).

The rest of the paper describes the system in more detail, focussing
on the question of how rules and statistics are combined in the
translation component. Section~\ref{Section:SLT-system} describes the
overall architecture of SLT.  Section~\ref{Section:Examples} gives
examples of typical non-trivial translation problems from the
English/French ATIS domain, and the way they are dealt with.
Finally, Section~\ref{Section:Results} summarizes the current
implementation status of the project, and presents the results of
tests carried out on a recent version of the prototype.

\section{The SLT system}
\label{Section:SLT-system}

The SLT system consists of a set of individual processing modules,
linked together in a pipelined fashion. The input speech signal is
processed by the SRI DECIPHER(TM) recognizer \cite{Murveit:93}, and an
N-best list of hypotheses is passed to the source language processor,
a copy of the SRI Core Language Engine (CLE; \cite{CLE}) loaded with
an English grammar and lexicon. The CLE produces for each speech
hypothesis a set of possible analyses in Quasi Logical Form, and uses
trainable preference methods to select the most plausible hypothesis
and analysis
\cite{AlshawiCarter:94,HLT-Nbest}.

The QLF analysis selected as most plausible is passed to the transfer
component, which first annotates it with extra information in a
rule-based pre-transfer phase. Next, a set of possible target-language
QLFs is created, using the unification-based transfer rules
\cite{Transfer-ACL-91}. The target QLFs are
stored in a ``packed'' form \cite{Tomita:86} to avoid a combinatoric
explosion when many transfer choices are non-deterministic.  A
rule-based post-transfer phase then performs some simple rewriting of
the transferred QLFs, following which a second set of trained
statistical preferences extract the most plausible transferred QLF and
``unpack'' it into a normal representation. The selected
target-language QLF is passed to a second copy of the CLE, loaded with
a target-language grammar and lexicon, which generates a surface string
using the Semantic Head-Driven algorithm \cite{SHD}. Finally, the
target-language string is passed to a speech synthesizer and converted
into output speech.

Most of this processing has already been covered in detail in
\cite{SLT-report}, with reference to the Swedish version. The rest of
this section will describe the new functionalities added since then:
trainable transfer preferences, transfer packing, and the use of pre-
and post-transfer phases. The final sub-section briefly summarizes the
main features of the French language description.

\subsection{Trainable transfer preferences}
\label{Transfer-preferences}

The basic preference model and training method for transfer
preferences is the one described in \cite{AlshawiCarter:94} and
\cite{HLT-Nbest}, suitably adapted for the transfer task; a
brief summary follows.  We start with a training corpus, consisting of
a set of utterances, each paired with a list of possible output
sentences produced by the transfer component.  A human judge marks
each transfer as either acceptable or unacceptable. In line with
the noisy-channel statistical model of translation described in
\cite{Brown:90}, the plausibility of a new candidate transfer is now
defined to be a real number, calculated as a weighted sum of two
contributions: the {\it transfer rule score}, and the {\it target
language model score}. The first of these represents the relative
plausibility of the rules used to make the transfer, and the second
the plausibility of the target QLF produced.

The transfer rule score and the target language model score are
computed using the same method; for clarity, we first describe this
method with reference to transfer rules.  The transfer rule score for
the bag of transfer rules used to produce a given target QLF is a sum
of the {\it discriminant scores} for the individual transfer rules.
The discriminant score for a rule $R$ is calculated from the
training corpus, and summarizes the reliability of $R$ as an indicator
that the transfer is correct or incorrect. The intent is that transfer
rules which tend to occur more frequently in correct transfers than
incorrect ones will get positive scores; those which occur more
frequently in incorrect transfers than correct ones will get negative
scores.

More formally, we define the discriminant score for $R$, $d(R)$, as
follows. We find all possible 3-tuples $(S, T_1, T_2)$ in the
training corpus where
\begin{itemize}
\item $S$ is a source language utterance,
\item $T_1$ and $T_2$ are possible transfers for $S$, exactly one of
which is correct,
\item The transfer rule $R$ is used in exactly one of $T_1$ and $T_2$.
\end{itemize}
If $R$ occurs in the correct hypothesis of the pair $(T_1, T_2)$, we
call this a ``good'' occurrence of $R$; otherwise, it is a ``bad''
one.  Counting occurrences over the whole set, we let $g$ be the total
number of good occurrences of $R$, and $b$ be the total number of bad
occurrences. $d(R)$ is then defined as
\begin{eqnarray*}
d(R) =    \left\{
                 \begin{array}{ccl}
                  \log_2(2(g + 1)/(g + b + 2))  & if & g < b  \\
                                                0  & if & g = b \\
                  -\log_2(2(b + 1)/(g + b + 2)) & if & g > b
	         \end{array} \right.
\end{eqnarray*}
This formula is a symmetric, logarithmic transform of the function $(g
+ 1)/(g + b + 2)$, which is the expected {\it a posteriori}
probability that a new $(S, T_1, T_2)$ 3-tuple will be a good
occurrence of $R$, assuming that, prior to the quantities $g$ and $b$ being
known, this probability has a uniform {\it a priori} distribution on
the interval [0,1].

The target language model score is defined similarly. The first step
is to extract a bag of ``semantic triples'' \cite{AlshawiCarter:94}
from each possible transferred QLF in the training corpus, following
which each individual triple is assigned a discriminant score using
the method above. Semantic triples encode grammatical relationships
between head-words; we have generalized the original definition from
\cite{AlshawiCarter:94} to include relationships involving
determiners, since these are important for transfer. Thus for example
the normal reading of the English sentence
\begin{quote}
{\it Show flights with a stop.}
\end{quote}
would include the triples
\begin{verbatim}
(show,obj,flight)    (show,obj,bare_plur)
(bare_plur,det,flight) (flight,with,stop)
(flight,with,a)        (a,det,stop)
\end{verbatim}
In Section~\ref{Section:Examples} below, we will present examples
illustrating how the two components of the transfer preference model
combine to solve some non-trivial transfer problems.

\subsection{Pre- and post-transfer}

Ideally, we would like to say that unification-based rules
and trainable transfer preferences constituted the whole transfer
mechanism. In fact, we have found it necessary to bracket the
unification-based transfer component between pre- and post-transfer
phases. Each phase consists of a small set of rewriting rules, which
are applied recursively to the QLF structure. It would in principle
have been possible to express these as normal unification-based
transfer rules, but efficiency considerations and lack of
implementation time persuaded us to adopt the current solution.

The pre-transfer phase implements a simple treatment of reference
resolution or coercion, which at present only deals with a few cases
important in the ATIS domain. Most importantly, QLF constructs
representing bare code expressions used as NPs are annotated with the
type of object the code refers to. Code expression are frequent
in ATIS, and the type of referent is always apparent from the code's
syntactic structure. The extra information is necessary to obtain a
good French translation: flight codes must be prefaced with
{\it le vol} (e.g. {\it C O one three three $\rightarrow$ le vol C O
cent trente-trois}) while other codes are translated literally.

The post-transfer phase reduces the transferred QLF to a canonical
form; the only non-trivial aspect of this process concerns the
treatment of nominal and verbal PP modifiers. In French, PP modifier
sequences are subject to a strong ordering constraint: locative PPs
should normally be first and temporal PPs last, with other PPs in
between. In the limited context of the ATIS domain, this requirement
can be implemented fairly robustly with a half-dozen simple rules, and
leads to a marked improvement in the quality of the translation.

\subsection{Transfer packing}

As already indicated, the basic philosophy of the transfer component is
to make the transfer rules more or less context-independent, and let
the results be filtered through the statistically trained transfer
preferences. The positive side of this is that the transfer rules are
robust and simple to understand and maintain. The negative side is
that non-deterministic transfer choices multiply out, giving a
combinatoric explosion in the number of possible transferred QLFs.

To alleviate this problem, transferred QLFs are {\it packed}, in
the sense of \cite{Tomita:86}; lexical transfer ambiguity is left
``unexpanded'', as a locally ambiguous structure in the target QLF.
It is possible to compute preference scores efficiently on the
packed QLFs, and only unpack the highest-scoring candidates; this
keeps the transfer phase acceptably efficient even when several
thousand transferred QLFs are produced.

The following example illustrates how transfer packing works. The
source utterance is
\begin{quote}
{\it flights on Monday}
\end{quote}
and the packed transferred QLF (in slightly simplified form) is:
\begin{verbatim}
elliptical_np(
  term(/|\(1,[def_plur,
              indef_plural,
              bare_plur]),
       C^[and,
           [vol1,C],
           form(prep(/|\(2,[a_bord_de,
                            temporal_np,
                            sur,
                            pour,
                            avec])),
                term(/|\(3,[def_sing,
                            bare_sing])
                     E^[lundi1,E]))]))
\end{verbatim}
This contains three lexical transfer ambiguities, reflecting the
different ways of translating the bare singular and bare plural
determiners, and the preposition {\it ``on''}. In this case,
the transfer preferences determine that the best choices are
to realise English bare plural as French definite plural, English
bare singular as French definite singular, and {\it ``on''} as an
implicit temporal NP marker. Substituting these in, the preferred
unpacked QLF is
\begin{verbatim}
elliptical_np(
  term(def_plur,
       C^[and,
           [vol1,C],
           form(temporal_np,
                term(def_sing,
                     E^[lundi1,E]))]))
\end{verbatim}
producing the French surface output
\begin{quote}
{\it les vols le lundi}
\end{quote}

\subsection{French language description}

The French language description is a straightforward adaptation of the
general unification-based CLE grammar and lexicon for English
\cite{CLE-grammar}. It covers most important French
constructions, including all those occurring frequently in the ATIS
domain. The most significant divergences compared to the English
language description are in the treatment of clitic pronouns, which
will be reported in detail elsewhere.  Very briefly, however, a
approach analogous to the standard idea of ``gap-threading'' has been
implemented, which uses difference lists to ``move'' the clitics from
their surface position next to the verb to their notional positions
(usually, but not necessarily, as verb complements).

A fairly complete treatment of French inflectional morphology has been
implemented, based on \cite{French-morphology}.
The French lexicon currently
contains about 750 stem entries (excluding proper nouns), which is adequate
to provide good coverage of the ATIS domain in the English to French
direction. Of these entries, about half are for function words and the
remainder for content words.

\section{Examples of non-trivial translation problems}
\label{Section:Examples}

This section will give examples of non-trivial translation problems
from the ATIS domain, and describe how SLT deals with them. We were
interested to discover that even a domain as simple as ATIS actually
contains many quite difficult transfer problems; also, that
English/French is considerably more challenging than the
English/Swedish transfer pair used in the original SLT system.  We
will begin by giving examples\footnote{All examples presented in this
section are correctly processed by the current French version of SLT.}
where it is fairly clear that the problem is essentially grammatical
in nature, and thus primarily involves the rule-based part of the
system; later, we give examples where the problem mainly involves the
preference component, and examples where both types of knowledge are
needed.

An obvious case of a grammatical phenomenon is agreement, which is
considerably more important in French than in English; the rules for
agreement are rigid and well-defined, and easy to code in a
feature-based formalism. Quite frequently, however, they relate words
which are widely separated in the surface structure, which makes them
hard to learn for surface-oriented statistical models. For example,
there are many instances in ATIS of nouns which in French are
postmodified both by a PP and by a relative clause, e.g.
\begin{quote}
{\it Flights from Boston to Atlanta leaving before twelve a m \\
     $\rightarrow$ Les vols de Boston \`{a} Atlanta qui partent avant midi}
\end{quote}
Here, the verb {\it partent} has to agree in number and person with
the head noun {\it vols}, despite the gap of five surface words
in between.

Many problems related to word-order also fall under the same heading,
in particular those relating to question-formation and the position of
clitic pronouns. For example, French YN-questions can be formed in
three ways: by inversion of subject and main verb, by prefacing the
declarative version of the clause with the question particle {\it
est-ce que}, or by ``complex inversion'', fronting the subject and
inserting a dummy pronoun after the inverted verb.  If the subject is
a pronoun, only the first and second alternatives are allowed; if it
is {\it not} a pronoun, only the second and third are valid.  Thus for
example
\begin{quote}
{\it Does it leave after five p m? \\
     $\rightarrow$ Part-il apr\`{e}s dix-sept heures? \\
     $\rightarrow$ Est-ce qu'il part apr\`{e}s dix-sept heures? \\
     $\rightarrow$ *Il part-il apr\`{e}s dix-sept heures? \\
                                              \\
     Does that flight serve meals?\\
     $\rightarrow$ *Sert ce vol des repas? \\
     $\rightarrow$  Est-ce que ce vol sert des repas? \\
     $\rightarrow$ Ce vol sert-il des repas? }
\end{quote}
Embedded questions constitute another good example of a mainly
grammatical problem. Just as in English, French embedded
questions normally have the uninverted word-order, e.g.
\begin{quote}
{\it Tell me {\bf when these flights arrive in Boston}\\
     $\rightarrow$ Dites-moi {\bf quand ces vols arrivent \`{a} Boston}}
\end{quote}
However, if the main verb is {\it \^{e}tre} with an NP complement, the
inverted word-order is obligatory, e.g.
\begin{quote}
{\it Tell me {\bf what the cheapest fares are}\\
     $\rightarrow$ Dites-moi {\bf quels sont les tarifs les moins chers}\\
     $\rightarrow$ *Dites-moi {\bf quels les tarifs les moins chers sont}}
\end{quote}
In ATIS, embedded questions occur in about 1\% of all corpus
sentences; this makes them too frequent to ignore, but rare
enough that a pure statistical model will probably have difficulties
finding enough training examples to acquire the appropriate
regularities. The relevant facts are however quite easy to state as
grammatical rules. Moreover, they are domain-independent, and can thus
be reused in different applications.

In contrast, there are many phenomena, especially involving
word-choice, which are hard to code as rules and largely domain- and
application-dependent. As mentioned earlier in
Section~\ref{Section:Introduction}, the translation of prepositions
and determiners is most frequently determined on
collocational grounds; in our framework, this means that the
information used to decide on an appropriate translation is
primarily supplied by the transfer preferences. We will now
decribe in more detail how the idea works in practice.

Recall that the preference score for a given transfer candidate is a
weighted sum of a channel contribution (discriminants on transfer
rules) and a target language model score (discriminants from target
language semantic triples). The transfer rule discriminants make
transfer rules act more or less strongly as defaults. If a transfer
rule $R$ is correct more often than not when a choice arises, it will have
a positive discriminant, and will thus be preferred if there is no
reason to avoid it. If use of $R$ produces a strong negative
target-language discriminant, however, the default will be overridden.

Let us look at some simple examples. The English indefinite singular
article {\it ``a''} can be translated in several ways in French, but
most often it is correct to realise it as an indefinite singular ({\it
``un''} or {\it ``une''}). The discriminant associated with the
transfer rule that takes indefinite singular to indefinite singular is
thus fairly strongly positive. There are however several French
prepositions which have a strong preference for a bare singular
argument; for instance, {\it ``flights without a stop''} is almost
always better translated as {\it ``les vols sans escale''} than {\it
``les vols sans une escale''}.  In cases like these, the {\it
a}-to-{\it un} rule will be wrong, and the less common rule that takes
indefinite singular to bare singular will be right. So if enough
training examples are available, the negative discriminant associated
with the semantic triple
\begin{verbatim}
(vol, sans, indef_sing)
\end{verbatim}
will have a higher absolute value than the positive discriminant
associated with {\it a}-to-{\it un}, and can overrule it.

Similar considerations apply to prepositions. In the ATIS domain, most
prepositions have several possible translations, none of which are
strongly preferred. For example, the channel score discriminants
associated with the transfer rules {\it on}-to-{\it sur} and
{\it on}-to-{\it avec} both have low absolute values; the first
is slightly negative, and the second slightly positive.
Target language triples associated with these prepositions are however
in general more definite: the triples
\begin{verbatim}
(aller avec <airline>)
(renseignement sur transports)
\end{verbatim}
are both strongly positive, while
\begin{verbatim}
(aller sur <airline>)
(renseignement avec transports)
\end{verbatim}
are strongly negative. The net result is that the target language
contribution makes the decision, and as desired we get {\it ``fly on
Delta''} and {\it ``information on flights''} going to {\it ``aller
avec Delta''} and {\it ``des renseignements sur les vols''} rather
than {\it ``aller sur ...''} and {\it ``des renseignements avec
...''}.

In general, a combination of rules and collocational information is
needed to translate a construction. A good example is the English
implicit singular mass determiner, which is common in ATIS.
Grammatical rules are used to decide that there is a singular mass
determiner present, following which the correct translation is
selected on collocational grounds. An elementary French grammar will
probably say that the normal translation should either be the French
partitive singular determiner, e.g.
\begin{quote}
{\it I drink {\bf milk} \\
      $\rightarrow$ Je bois {\bf du lait}}
\end{quote}
or else the definite singular, e.g.
\begin{quote}
{\it I like {\bf cheese} \\
      $\rightarrow$ J'aime {\bf le fromage}}
\end{quote}
In the ATIS domain, it happens that the nouns which most frequently
occur with mass singular determiner are {\it ``transportation''} and
{\it ``information''}, both of which are conventionally singular in English
but plural in French. Because of this, neither of the standard rules
for translating mass singular gets a strong positive discriminant
score, and once again the target language model tends to make the decision.
For instance, if the head noun is {\it ``transportation''}, it is most
often correct to translate the mass singular determiner as a definite
plural, e.g.
\begin{quote}
{\it Show me {\bf transportation} for Boston\\
      $\rightarrow$ Indiquez-moi {\bf les transports} pour Boston}
\end{quote}
This is captured in a strong positive discriminant score associated
with the target language triple
\begin{verbatim}
(def_plur, det, transport)
\end{verbatim}
Note that the translation {\it ``transportation''} to {\it ``les
transports''} is only a preference, not a hard rule; it can be
overridden by an even stronger preference, such as the preference against
having a definite plural subject of an existential construction. So we
have e.g.
\begin{quote}
{\it Is there {\bf transportation} in Boston?\\
      $\rightarrow$ Y a-t-il {\bf des transports} \`{a} Boston?\\
      $\rightarrow$ *Y a-t-il {\bf les transports} \`{a} Boston?}
\end{quote}

\section{Implementation status and results}
\label{Section:Results}

So far, the French version of SLT has consumed about eleven
person-months of effort over and above the effort expended on the
original English/Swedish SLT project. Of this, about seven
person-months were spent on the French language description, two on
transfer rules, and two on other tasks. The small quantity of effort
required to develop a good French language description underlines the
extent to which its structure overlaps with that of the original
English grammar and lexicon.

We now describe preliminary experiments designed to test the
performance of the system. A set of 2000 ATIS utterances was used,
randomly selected from the subset of the ATIS corpus consisting of A
or D class\footnote{This means roughly that the sentence represented a
valid inquiry to the database, either alone or in the context in which
it was uttered.} utterances of length up to 15 words, which had not
previously been examined during the development of the French version
of SLT.  Utterances were supplied in text form, i.e. the speech
recognition part of the system was not tested here.

Each utterance was analysed using the English language version of the
CLE, and for the 1847 sentences where at least one QLF was produced the
most plausible QLF was selected using the preference methods described
in \cite{AlshawiCarter:94}.  This was then submitted to the transfer
phase, and a set of transfer candidates produced. A simple set of
hand-coded transfer preferences was applied, and one French surface
string was generated for each of the five highest-scoring transfer
candidates.  A native French speaker fluent in
English judged each generated string as being either an acceptable or
an unacceptable translation of the source utterance. Translations were
only regarded as acceptable if they were fully grammatical, preserved
the meaning of the source utterance, and used a stylistically natural
choice of words. The judging process took approximately eight hours,
averaging three seconds per source/target pair.

The annotated N-best transfer corpus was then used to train a new set
of preferences using the method described in
Section~\ref{Transfer-preferences}; the corpus was divided into five
equal pieces, each fifth being held out in turn as test data with the
remaining four-fifths used as training. Finally, the derived
preferences were tested for accuracy. Of the 1847 transfer sets, there
were 1374 for which at least one acceptable transfer was in the top
five candidates\footnote{There were a further 246 sets in which at
least one candidate translation was produced; in most of these cases,
the best translation was comprehensible and grammatically correct, but
was rejected on stylistic grounds.}. The trained transfer preferences
selected an acceptable candidate in 1248 of these 1374 cases (91\%);
in contrast, random choice among the top five gave a baseline score of
826 acceptable transfers, or 60\%. We regard this as a promising
initial result, and intend soon to repeat the experiment with a larger
set of 5000--10000 sentences. We also anticipate significant
improvements over the next few months from planned extensions and
refinements to the French language description.

\end{document}